\documentclass{article}

\usepackage{microtype}
\usepackage{graphicx}
\usepackage{subfigure}
\usepackage{booktabs} % for professional tables

\usepackage{dsfont}
\usepackage{amsmath,amssymb}

\usepackage[makeroom]{cancel}

\usepackage{hyperref}

\usepackage[accepted]{icml2020}

\DeclareUnicodeCharacter{0308}{ }

\graphicspath{{figures/}}

\begin{document}

\twocolumn[
\icmltitle{Clustering Survival Data using a Mixture of Non-parametric Experts}



\icmlsetsymbol{equal}{*}

\begin{icmlauthorlist}
\icmlauthor{Gabriel Buginga}{ufrj}
\icmlauthor{Edmundo A. de Souza e Silva}{ufrj}
\end{icmlauthorlist}

\icmlaffiliation{ufrj}{PESC, Federal University of Rio de Janeiro, Rio de Janeiro, Brazil}

\icmlcorrespondingauthor{Gabriel Buginga}{buginga@cos.ufrj.br}
\icmlcorrespondingauthor{Edmundo de Souza e Silva}{edmundo@land.ufrj.br}

\icmlkeywords{Machine Learning, Survival Analysis}

\vskip 0.3in
]




% This is the name of the Appendix or Supplementary Material, depending on which Journal it's going to be published. The second definition is the name of the model proposed.
\newcommand{\nameAppendix}{appendix }
\newcommand{\nameModel}{\textbf{SurvMixClust} }

\begin{abstract}
Survival analysis aims to predict the timing of future events across various fields, from medical outcomes to customer churn. However, the integration of clustering into survival analysis, particularly for precision medicine, remains underexplored. This study introduces SurvMixClust, a novel algorithm for survival analysis that integrates clustering with survival function prediction within a unified framework. SurvMixClust learns latent representations for clustering while also predicting individual survival functions using a mixture of non-parametric experts. Our evaluations on five public datasets show that SurvMixClust creates balanced clusters with distinct survival curves, outperforms clustering baselines, and competes with non-clustering survival models in predictive accuracy, as measured by the time-dependent c-index and log-rank metrics. 
\end{abstract}

\section{Introduction}

Survival analysis, also known as time-to-event analysis, involves predicting the likelihood of an event occurring, such as death, product lifetime, customer churn, and illness remission \cite{fleming2011counting, Kleinbaum2010}. This requires using probabilistic models to stratify risk profiles, determine individual event distribution, and identify the impact of features on event time. Numerous methods have been developed for this purpose, including the nonparametric Kaplan-Meier estimator used to generate a marginal survival function \cite{kaplan1958nonparametric}, and the Cox Proportional Hazards, a semi-parametric linear hazards method \cite{cox1972regression}. Parametric methods, such as Weibull, exponential, and log-normal distributions, are also utilized along with machine learning models such as survival tree \cite{leblanc1993survival}, random survival forest \cite{ishwaran2008random}, and survival support vector machine \cite{polsterl2015fast}, which have shown very good performance \cite{wang2019machine}. Deep learning has also contributed to survival analysis by extracting complex interactions between features and the time-to-event process. A notable early example is the DeepSurv model's extension of CoxPH \cite{katzman2018deepsurv}.

Cluster analysis of survival data can identify similar groups in time-to-event distribution, aiding in disease subtyping, risk stratification, and clinical decision-making. Precision medicine can benefit from these algorithms \cite{collins2015new}. Currently, only a few specialized algorithms exist for clustering individuals based on their survival function, yet this topic is beginning to gain increased attention.

Some models attempt to create clusters without jointly learning clustering and prediction. For instance, they achieve this by employing a pipeline approach, with the aid of a Cox Proportional Hazards (CoxPH) model \cite{tosado_clustering_2020} or a hierarchical clustering method \cite{chen_algorithm_2016, ahlqvist2018novel}. Since these algorithms aim to optimize clustering, they may not be effective in predicting survivability with reasonable accuracy. To address this issue, models that jointly learn clustering and prediction have been proposed recently~\cite{chapfuwa2020survival,manduchi2021deep}. 
These works demonstrate that the cluster they found and the corresponding predictive performance surpass those of the previous algorithms mentioned earlier.
Nevertheless, our experiments indicate that their predictive performance can still be improved.

\textbf{Contributions}. Our contributions are summarized below:
\begin{itemize}
    \item We propose a new algorithm named \nameModel for clustering survival data. It has the ability to learn a latent representation for clustering while also learning a survival function for each cluster. Similarly to \cite{chapfuwa2020survival}, the algorithm can generate a custom survival function for each data point.
    
    \item The \nameModel algorithm can identify clusters with highly diversified survival functions, each displaying distinctive curves. Additionally, it can find clusters with a balanced number of data points. When compared to other clustering algorithms, it has shown superior predictive performance, as evaluated using the time-dependent c-index metric \cite{Antolini2005}, across all datasets. Additionally, our algorithm outperformed others in three out of five datasets when assessed using the log-rank metric \cite{Mantel1966}.
    
    \item \nameModel exhibits competitive predictive performance when compared to non-clustering survival models, such as the Random Survival Forest model. This comparison is quantified using the time-dependent c-index survival metric \cite{Antolini2005}.
    
    \item The code for the model has been made publicly available and can be found at \href{https://github.com/buginga/SurvMixClust}{https://github.com/buginga/SurvMixClust}. We follow the basic scikit-learn API.
\end{itemize}

% Survival clusterization methods in the statistics literature mostly assume parametric models, as seen in \cite{lachos_finite_2017} or \cite{hanson_modeling_2006}.

% Experimental evidence demonstrates that our solution delivers competitive predictive performance, highly distinctive and informative clusters, and interpretable results.

% Our contribution to this problem is to build a finite mixture of nonparametric distributions, using the features to model the mixture proportions with a multinomial logistic regression. In other words, we propose a mixture of experts with a gating network for the right-censored time-to-event clusterization problem. It's trained via maximum likelihood with a variety of the Expectation-Maximization (EM) algorithm. 

\section{The Model}

The standard survival analysis framework used throughout the paper is the right-censored formalization. Here,  $T^{*}$ represents the random variable indicating the ground-truth event time, and $C^{*}$ denotes the censoring time. Note that, in practice, we only observe samples from $T=\text{min}\{T^{*},C^{*}\}$ and $D=\mathds{1}\{T^{*}\leq C^{*}\}$, representing the minimum between the ground-truth event time and the censoring time, and the indicator function for the event occurrence, respectively.
Note that $t_i$ is a lower bound on the true event time when $d_i=0$. 
For instance, in a medical study where the event to be predicted is death, if a patient does not return for a follow-up to complete treatment, $d_i=0$, the last observed time is the censoring time, and the true time of death is evidently greater than $t_i$. Conversely, $d_i$=1 indicates that $t_i$ is the actual time of death.

We assume that data points originate from an i.i.d. process, with samples of the form \(\mathcal{D} = \{\boldsymbol{x}_{i}, \boldsymbol{y}_{i}\}_{i=1}^{n}\), where \(\boldsymbol{x}_{i} \in \mathbb{R}^{m}\) (with \(m\) representing the number of features), and \(\boldsymbol{y}_{i} = (t_{i}, d_{i})\), corresponding to \(T \sim t_{i}\), \(D \sim d_{i}\), and \(X \sim \boldsymbol{x}_{n}\). Here, \(Z\) denotes the discrete latent variable that models the clusters we aim to discover. Furthermore, we assume random censoring, implying that \(T^{*}\) is statistically independent of \(C^{*}\).

\begin{figure}[ht]
\vskip 0.2in
\begin{center}
\centerline{\includegraphics[width=\columnwidth]{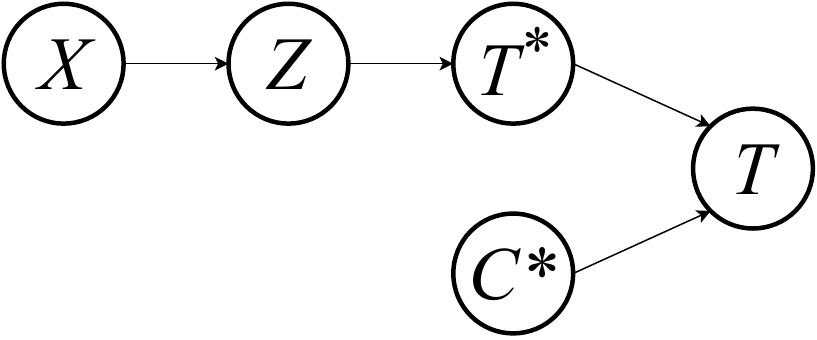}}
\caption{Graph model representing independence assumptions for the main model. Notice how the features $X$ can only influence $T^{*}$ via the clusterization $Z$.}
\label{fig:graph model}
\end{center}
\vskip -0.2in
\end{figure}

Figure \ref{fig:graph model} illustrates the graph model resulting from the independence assumptions. These assumptions are consistent with our reliance on the Kaplan-Meier estimator \cite{kaplan1958nonparametric} and represent a modeling choice that is simple and aids interpretability \cite{bouveyron2019model}. It is important to note that the features, denoted as $X$,  influence $T^{*}$ solely through the clusterization process, represented by $Z$.
The features of each data point influence $T^{*}$  through the use of multinomial logistic regression functions as models.

The iterative training process to discover clusters can be succinctly described as follows.
In a specific step of the process, data points are assigned to clusters based on the overall model derived from the previous step. Once all data points are clustered for the current step, new survival functions are computed using the Kaplan-Meier estimate. Furthermore, the parameters for the multinomial logistic regression functions are re-estimated. The following sections provide further details of the algorithm.
All relevant notation used throughout the paper is presented in Table \ref{table:notation}.

\begin{table}[t]
\caption{Notations and definitions used throughout the paper.}
\label{table:notation}
\vskip 0.15in
\begin{center}
\begin{small}

\begin{tabular}{l p{6cm}}
\toprule
Notation & Description\\
\midrule
$T^{*}$     & random variable for ground-truth time-to-event\\
$C^{*}$ 	& random variable for ground-truth censoring process\\
$f_{C^{*}}(t)$ & density function for $C^{*}$\\
$S_{C^{*}}(t)$ & survival function for $C^{*}$\\
$f(t)$ & density function for $T^{*}$\\
$S(t)$ & survival function for $T^{*}$\\
$T$ 		& random variable for possibly censored time-to-event, defined as $T=\text{min}\{T^{*},C^{*}\}$\\
$t_i$ & possibly censored time-to-event for datapoint $i$\\
$D$ & binary random variable for censoring indication; defined as $D=\mathds{1}\{T^{*}\leq C^{*}\}$, i.e. $D=1$ represents that the event happened, and $D=0$ that it was censored\\
$d_i$ & censoring indication for datapoint $i$\\
$Z$ & discrete random variable taking values in $\{1,...,K\}$, modelling cluster attribution\\
$\boldsymbol{x}_{i}$ & $m$-dimensional vector of covariates for datapoint $i$\\
$\boldsymbol{y}_{i}$ & 2-dimensional label vector for datapoint $i$, defined as $\boldsymbol{y}_{i}=(t_{i}, d_{i})$\\
$\tau$ & multinomial logistic regression modeling the mixture proportions\\
$K$ & integer indicating the number of clusters for the main model\\
$\boldsymbol{\theta}$ & includes all parameters for the EM  \\
$\theta_{k}$ & indicates the non-parametric distribution for cluster $k$\\
$\boldsymbol{\beta}$ & K-dimensional vector of the coefficients for the multinomial regression\\
\bottomrule
\end{tabular}

\end{small}
\end{center}
\vskip -0.1in
\end{table}

\subsection{Definition}

The model is a finite mixture of $K$ nonparametric distributions, wherein the mixing weights are calculated via a multinomial logistic regression using the features as input. Recall from Figure \ref{fig:graph model} that $X$ only influences $T^{*}$ through $Z$. Additionally, each cluster possesses its own fixed nonparametric distribution, which models how  $T^{*}$ behaves.   

Logistic regression was selected because it's one of the most simple and interpretable models that can be used to model mixing proportions \cite{bouveyron2019model}. When using features, the model-based clustering approach often selects logistic regression as the primary option \cite{bouveyron2019model}. Based on the assumption that only the features $X$ are utilized to select the mixing proportions, It is enough only to inspect how logistic regression is making inferences to understand how cluster assignments are determined for each data point. Expectation-maximization training can require dozens of iterations, making a lightweight model desirable. Even so, as Section \ref{expectation} makes clear, the logistic regression model can be easily exchanged by any other probabilistic model of a finite discrete random variable. This change can be helpful for datasets containing features with an exploitable structure, like images, for which a convolutional neural network might be a better fit.

Consider $\theta_{k}$ indicating the non-parametric distribution for cluster $k$, let $f(t)=P(T^{*}=t)$ and $S(t)=P(T^{*}>t)$. The model with $K$ clusters can then be presented as:

\begin{equation}
\label{eq:model density}
f\left({t}_{i}^{*} \mid \boldsymbol{x}_{i}\right)=\sum_{k=1}^{K} \tau_{k}\left(\boldsymbol{x}_{i}\right) f\left({t}_{i}^{*} \mid \theta_{k}\right)
\end{equation}

\begin{equation}
\tau_{k}\left(\boldsymbol{x}_{i}\right)=\frac{\exp \left(\beta_{k}^{T} \boldsymbol{x}_{i}\right)}{\sum_{l=1}^{K} \exp \left(\beta_{l}^{T} \boldsymbol{x}_{i}\right)}
\end{equation}

It is possible to obtain a model form that includes the survival function, which is more commonly used in the literature of survival analysis:

\begin{equation}
\label{eq:survival function}
S\left({t}_{i}^{*} \mid \boldsymbol{x}_{i}\right)=\sum_{k=1}^{K} \tau_{k}\left(\boldsymbol{x}_{i}\right) S\left({t}_{i}^{*} \mid \theta_{k}\right)
\end{equation}

The distributions and parameters to be found are $\boldsymbol{\theta}=\left\{S(.|\theta_{z}), f(.|\theta_{z})\right\}^{K}_{z=1}\cup\left\{\boldsymbol{\beta}\right\}$.

\subsection{Training with Expectation-Maximization}
\label{expectation}

%\textbf{There is an important source of confusion here.} Like most models in survival analysis, the model \ref{eq:model density} models $T^{*}$, It does \textbf{NOT} model $T=\text{min}\{T^{*},C^{*}\}$. Remember that survival data provides $T$ and $D$, i.e., possibly censored event time and binary variable showing censoring status; while we want to model $T^{*}$, the non-censored event time. Therefore, when writing the complete data log-likelihood for the EM, we start from a harder position than if we were dealing with non-censored data. More specifically, we start trying to increase the likelihood of $\mathcal{D}^{l}=\left\{\boldsymbol{x}_{i},\boldsymbol{y}_{i}, z_{i}\right\}_{i=1}^{n}=\left\{\boldsymbol{x}_{i}, t_{i}, d_{i}, z_{i}\right\}_{i=1}^{n}$. Where $\mathcal{D}^{l}$ differs from $\mathcal{D}$ by including information from the not-seen latent variable $Z$. The problem is that the main model's $T^{*}$ appears only implicitly. A non-censored dataset would have been $\mathcal{D}^{\mathrm{non-censored}}=\left\{\boldsymbol{x}_{i},\boldsymbol{y}_{i}, z_{i}\right\}_{i=1}^{n}=\left\{\boldsymbol{x}_{i}, t_{i}^{*}, z_{i}\right\}_{i=1}^{n}$, then there is no problem and we can apply EM directly. That's not the case with censored survival data. Fortunately, It just so happens that we're able to use a model of $T^{*}$ to increase the likelihood of $T$ and $D$.

The training is done via a standard combination of maximum likelihood estimation and the Expectation-Maximization \cite{dempster1977maximum} bound optimization algorithm. We first need to derive the expected complete data log-likelihood equation as a function of the data $\mathcal{D}=\left\{\boldsymbol{x}_{i},\boldsymbol{y}_{i}\right\}_{i=1}^{n}$ and $\boldsymbol{\theta}$. Then, calculate the algorithmic details of the expectation and maximization steps.

Before obtaining the maximum likelihood estimator, we need a helpful formula from \cite{kvamme2019continuous}, derivation at \ref{derivations}. Noting that $S_{C^{*}}(t)=P(C^{*}>t)$ and $f_{C^{*}}=P(C^{*}=t)$, we can write:

\begin{equation}
\label{eq:trick1}
\begin{aligned}
&\mathrm{P}(T=t, D=d)=\\
&=\left[f(t)^{\mathbb{I}\left(d=1\right)} S(t)^{\mathbb{I}\left(d=0\right)}\right] \cdot\\ 
&\qquad \qquad \cdot \left[f_{C^{*}}(t)^{\mathbb{I}\left(d=0\right)}\left(S_{C^{*}}(t)+f_{C^{*}}(t)\right)^{\mathbb{I}\left(d=1\right)}\right]\\
\end{aligned}
\end{equation}

Equation \ref{eq:trick1} separates how $T^{*}$ and $C^{*}$ relate to $\mathrm{P}(T=t, D=d)$. If a event happened, $d=1$, then $D=d=1$, $\mathbb{I}\left(d=1\right)=1$ and $\mathbb{I}\left(d=0\right)=0$, making $\mathrm{P}(T=t, D=1)=\left[f(t)^{1} S(t)^{0}\right]\left[f_{C^{*}}(t)^{0}\left(S_{C^{*}}(t)+f_{C^{*}}(t)\right)^{1}\right]$$=\left[f(t)\right]\left[\left(S_{C^{*}}(t)+f_{C^{*}}(t)\right)\right]=\mathrm{P}\left(T^{*}=t, C^{*} \geq t\right)$. Similarly for the case $d=0$, $\mathrm{P}(T=t, D=0)=\mathrm{P}\left(T^{*}>t, C^{*}=t\right)$. 

%=\left[f(t)^{0} S(t)^{1}\right]\left[f_{C^{*}}(t)^{1}\left(S_{C^{*}}(t)+f_{C^{*}}(t)\right)^{0}\right]

%We are not parametrizing $f_{C^{*}}$ and $S_{C^{*}}$ (as an assumption). But, $f$ and $S$ are, in fact, parametrized as $f\left({t}_{i}^{*} \mid \theta_{k}\right)$ and $S\left({t}_{i}^{*} \mid \theta_{k}\right)$ .

Let's define $h(t)$ as follows: 
$h(t)=\left[f_{C^{*}}(t)^{\mathbb{I}\left(d=0\right)}\left(S_{C^{*}}(t)+f_{C^{*}}(t)\right)^{\mathbb{I}\left(d=1\right)}\right]$. 
Based on the assumptions presented in Figure \ref{fig:graph model}, $C^{*}$ and $Z$ are independent when compared directly. Therefore, we can recognize that $h(t_{n}|z_{n}=l)$ is equal to $h(t_{n})$. Then,

\begin{equation}
\label{eq:trick2}
\begin{aligned}
 \mathrm{P}&(T=t, D=d, Z=z \mid \boldsymbol{\theta}) =\\
& =\mathrm{P}_{\boldsymbol{\theta}}(Z=z) f(T=t, D=d \mid Z=z ; \boldsymbol{\theta})\\
& =\mathrm{P}_{\boldsymbol{\theta}}(Z=z){f\left(t \mid \theta_{z}\right)}^{\mathbb{I}\left(d=1\right)}{S\left(t \mid \theta_{z}\right)}^{\mathbb{I}\left(d=0\right)}h(t)\\
\end{aligned}
\end{equation}

%This procedure is almost the same as the one contained in section 3.1 in \cite{bordes2016stochastic}. 

Equation \ref{eq:trick2} will be used to calculate the expected complete data log-likelihood $L L^{t}(\boldsymbol{\theta})$. We use the following notation for the cluster's labels: $z_{n}\in \left[1,...K\right]$, and $z_{n k}=\mathbb{I}\left(z_{n}=k\right)$. For the Expectation step at the iteration $(t)$, it's necessary to calculate the posterior probability of the datapoint $n$ belonging to cluster $k$. The details are included in Appendix \ref{derivations}:

\begin{equation}
\label{eq:expectation}
\begin{aligned}
r_{nk}^{(t)}&=\mathrm{P}\left(z_{n}=k\mid t_{n}, d_{n}, x_{n}; \boldsymbol{\theta}^{(t)}\right)\\
&=\frac{\left[{f\left(t_{n} \mid \theta_{k}\right)}^{\mathbb{I}\left(d_{n}=1\right)}{S\left(t_{n} \mid \theta_{k}\right)}^{\mathbb{I}\left(d_{n}=0\right)}\right]\tau_{k}^{(t)}\left(\boldsymbol{x}_{i}\right)}{\sum_{l=1}^{K} \left[{f\left(t_{n} \mid \theta_{l}\right)}^{\mathbb{I}\left(d_{n}=1\right)}{S\left(t_{n} \mid \theta_{l}\right)}^{\mathbb{I}\left(d_{n}=0\right)}\right]\tau_{l}^{(t)}\left(\boldsymbol{x}_{i}\right)}\\
\end{aligned}
\end{equation}

Using the equation $\mathbb{E}\left[z_{nk}\right]=r_{nk}^{(t)}$ and equation \ref{eq:trick2}, the Maximization step at $(t)$ can be expressed as follows:

\begin{equation}
\label{eq:LL}
\begin{aligned}
& L L^{t}(\boldsymbol{\theta})=\\
& \sum_{n} \mathbb{E}_{q_{n}^{t}\left({z}_{n}\right)}\left[\log \mathrm{P}(T=t_{n}, D=d_{n}, Z=z_{n} \mid X=x_{n}; \boldsymbol{\theta}) \right]\\
&= \sum_{n} \sum_{k} r_{nk}^{(t)}\log \tau_{k}\left(\boldsymbol{x}_{n}\right) + \\
&\qquad + \sum_{n} \sum_{k} r_{nk}^{(t)} \mathbb{I}\left(d_{n}=1\right)\log {f\left(t_{n} \mid \theta_{k}^{(t)}\right)} +\\
&\qquad + \sum_{n} \sum_{k} r_{nk}^{(t)}\mathbb{I}\left(d_{n}=0\right)\log {S\left(t_{n} \mid \theta_{k}^{(t)}\right)} +\\
& \qquad + \sum_{n} \sum_{k} r_{nk}^{(t)} \left(h(t_{n})\right)\\
\end{aligned}
\end{equation}

As $\sum_{n,k} r_{nk}^{(t)} h(t)$ does not depend on $\boldsymbol{\theta}$, we will only maximize the remaining three terms of equation \ref{eq:LL} in order to increase the model likelihood. During the Expectation Maximization training, we calculate the likelihood of each data point belonging to each cluster based on the current parameters of the clusters. This is done in the expectation step, where we consider the survival curves ($S\left(t_{i} \mid \theta_{k}\right)$) associated with each cluster and the mixing proportions ($\tau_{k}$) provided by the logistic regression. Using these parameters, we calculate the posterior probability of the datapoint $n$ belonging to cluster $k$. Note that, even when logistic regression suggests a high probability of a data point belonging to a specific cluster, it might be reassigned to another cluster depending on the values of $S\left(t_{i} \mid \theta_{k}\right)$. Therefore, based on these probabilities, we attribute each point to the cluster with the highest posterior probability, i.e., we do a hard assignment. 

In the maximization step, we refine the parameters of each cluster to better fit the data. We adjust the clusters based on the current responsibilities assigned in the E-step, aiming to position the clusters so that they best explain the data points they are most likely to represent. To do this, we reestimate each $S\left(t_{i} \mid \theta_{k}\right)$ with the Kaplan-Meier estimator. We also retrain the logistic regression to predict the newly assigned clusters. We repeat the E-step and M-step until the clusterization stabilizes. After fitting the data, we obtain a model in the form of equation \ref{eq:survival function}. Each cluster's survival function is linked to the corresponding point in the feature space through logistic regression. The logistic regression assigns a particular cluster $k$ to each region, and within each region, the cluster's survival function $S\left(t_{i} \mid \theta_{k}\right)$ provides support for the points belonging to that cluster, increasing their likelihood.

\subsection{Training Algorithm}
Utilizing equations \ref{eq:LL} and \ref{eq:expectation}, the model is trained with \textit{stochastic} expectation maximization \cite{celeux1996stochastic}, similar to the one used in section 4.2 of \cite{bordes2016stochastic}. A single run of the entire algorithm follows the steps:

\begin{enumerate}
\setcounter{enumi}{-1}
\item \textbf{Hyperparameters}: establish the value of the number $K$ of clusters ($K \in \{2, 3, 4, ...\}$).

\item \textbf{Initialization}: For each $i \in \{1, ..., n\}$, $z_{i}$ is assigned a label from $\{1,...,K\}$ with equal probability, i.e., completely random assignment.
\end{enumerate}

Then, repeat until convergence the following two steps:

\begin{enumerate}
\setcounter{enumi}{1}
\item \textbf{E-Step}: For each $i \in \{1,...,n\}$:
\begin{equation}
\begin{aligned}
&q^{(t)}_{i}=\\
&=\arg \max _{k}\left({f\left(t_{i} \mid \theta_{k}^{(t)}\right)}^{\mathbb{I}\left(d=1\right)}{S\left(t_{i} \mid \theta_{k}^{(t)}\right)}^{\mathbb{I}\left(d=0\right)}\right.\cdot\\
&\qquad \qquad \qquad \qquad \qquad \cdot \left.\tau_{k}^{(t)}\left(\boldsymbol{x}_{i}\right)\right)\\
&r_{iq^{(t)}_{i}}^{(t)} = 1\\
&r_{ik}^{(t)} = 0, \quad \forall k \in \{1,...,K\} \: \mathrm{and} \: k\neq q^{(t)}_{i} \\
\end{aligned}
\end{equation}
\item \textbf{M-Step}: For each $k \in \{1,...K\}$, denote $\mathrm{cluster}_{k}=\{i \in \{1,...,n\}\: |\: r_{ik}^{(t)}=1\}$, i.e., $\mathrm{cluster}_{k}$ include all data points that were assigned the exclusive label of $k$. Repeat the following steps for each $k$.
\begin{enumerate}
\item $S\left(t \mid \theta_{k}^{(l+1)}\right)$ is estimated with the Kaplan-Meier estimator \cite{kaplan1958nonparametric} using the data points in  $\mathrm{cluster}_{k}$.
\item $f\left(t \mid \theta_{k}^{(l+1)}\right)$ is calculated using a nonparametric presmoothed estimator \cite{deUllibarri2013}, its bandwidth is selected via plug-in estimate and fixed for all EM steps in order to reduce computation time.
\item Finally, for estimating $\tau^{(l+1)}$ we train a multinomial logistic regression classifier. The labels are the $q^{(t)}_{i}$ calculated at the E-Step. Specifically, the training data for the classifier is $\mathcal{D}^{\mathrm{reg}}=\left\{\boldsymbol{x}_{i},q^{(t)}_{i}\right\}_{i=1}^{n}$. This training is done as a standard supervised learning problem.
\end{enumerate}
\end{enumerate}

\section{Experiments}

These experiments compare the proposed algorithm with other survival models, including algorithms that do not cluster and those that do. The former group includes the Random Survival Forest \cite{ishwaran2008random}, the Cox Proportional Hazards (CoxPH) \cite{cox1972regression}, Logistic Hazard (neural network model) \cite{kvamme2019continuous}. The latter group is comprised of our proposal, the Survival Cluster Algorithm (SCA) \cite{chapfuwa2020survival}, and \textbf{K-means Survival}.
For more details on methodology, including hyperparameters searched, refer to the \nameAppendix. The number of clusters searched for $K$ ranges from 2 to 7.

\textbf{K-means Survival} is a variant of the K-means algorithm that can be used as a survival model. It can provide a survival function for each data point by grouping the data into clusters through unsupervised training. The training process for this algorithm is identical to that of a standard K-means algorithm. It groups data into clusters by iteratively assigning data points to the nearest centroid and updating these centroids based on the mean of the points in each cluster, thereby minimizing the sum of squared distances between each data point and its corresponding cluster centroid. Notably, the algorithm does not rely on labels for training. The adaptation for survival problems only occurs during inference. When a patient's data point is assigned to a cluster, the algorithm generates a Kaplan-Meier curve for the population within that cluster. The algorithm then returns the Kaplan-Meier curve of the patient's assigned cluster as their final inferred survival function.

Table \ref{table:datasets} lists the publicly accessible datasets used for the experiments. SUPPORT, Study to Understand Prognoses Preferences Outcomes and Risks of Treatment \cite{Knaus1995}; FLCHAIN, The Assay of Serum Free Light Chain (FLCHAIN) \cite{Dispenzieri2012}; GBSG, The Rotterdam \& German Breast Cancer Study Group \cite{Foekens2000}; METABRIC, The Molecular Taxonomy of Breast Cancer International Consortium \cite{Curtis2012}; the Worcester Heart Attack Study (WHAS500), specifically the version with 500 patients \cite{Lemeshow2011}. In all experiments, continuous features were imputed with their mean and categorical features with their mode, followed by one-hot encoding. The labels were kept unchanged.  

The first metric utilized is the time-dependent C-index \cite{Antolini2005}, a modified version of Harrell's C-index \cite{Harrell1982}. It is a nonparametric statistic, ranging from 0 to 1, that measures the probability of two randomly chosen comparable pairs of patients being concordant according to model-given risks. The ``time-dependent'' part indicates that this risk is calculated using the full survival function, which changes with time. This metric is calculated by treating all models as purely time-to-event survival models. For an evaluation of clusterization, the logrank score between clusterized populations is used \cite{Mantel1966}.

\begin{figure*}[ht]
\vskip 0.2in
\begin{center}
\centerline{\includegraphics[width=\textwidth]{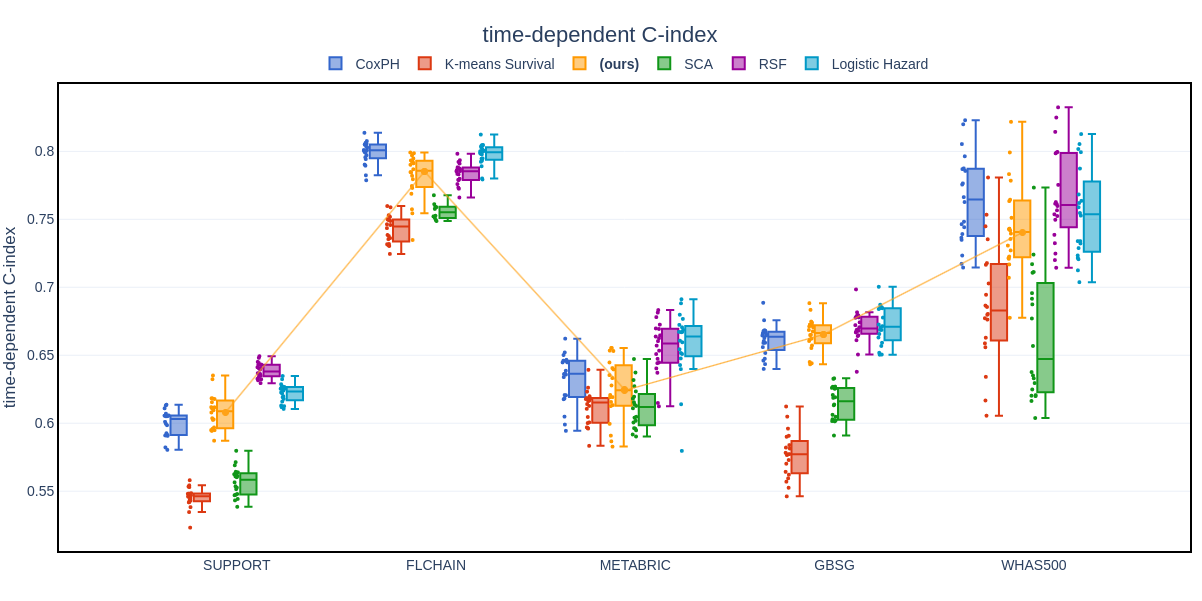}}
\caption{Time-dependent C-index across datasets and models. Each boxplot displays 20 samples.}
\label{fig:time-dependent cindex}
\end{center}
\vskip -0.2in
\end{figure*}

The chart shown in Figure \ref{fig:time-dependent cindex} demonstrates the time-dependent C-index results for different models and datasets. The approach proposed in this paper is represented by "\textbf{(ours)}" in the chart. The \nameModel \: model outperforms clustering-based models (SCA and \textbf{K-means Survival}) in all datasets. When compared to purely predictive models, \nameModel \: exhibits similar performance to the GBSG and WHAS500 datasets. However, the METABRIC and FLCHAIN datasets, fall into the second performance tier and are statistically similar to the RSF and CoxPH models, respectively. Based on the SUPPORT dataset, \nameModel \: ranks third in terms of performance, along with CoxPH.

\begin{figure*}[ht]
\vskip 0.2in
\begin{center}
\centerline{\includegraphics[width=\textwidth]{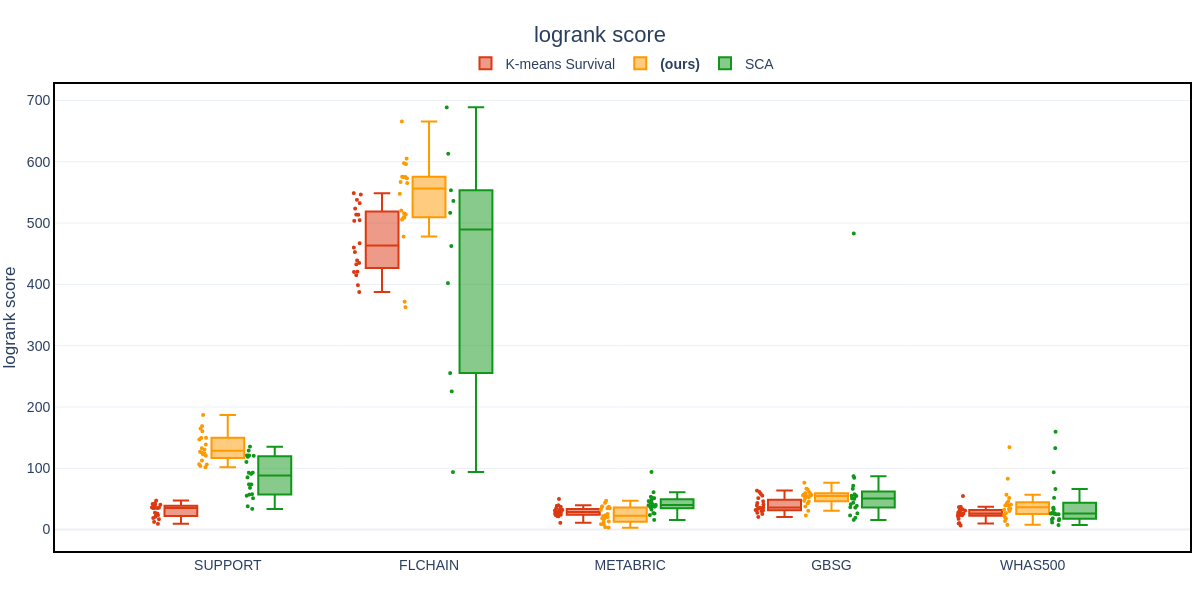}}
\caption{Logrank score across datasets and models. Each boxplot displays 20 samples.}
\label{fig:logrank}
\end{center}
\vskip -0.2in
\end{figure*}

Similarly, Figure \ref{fig:logrank} depicts the logrank results, which show higher log-rank metrics in the SUPPORT, FLCHAIN, and WHAS500 datasets. Additionally, the log-rank metric is similar to SCA for the GBSG dataset but worse for the METABRIC dataset. Note that the figure only includes models that cluster, as log-rank measures group differences in survival. 

\begin{figure*}[ht]
\vskip 0.2in
\begin{center}
\centerline{\includegraphics[width=\textwidth]{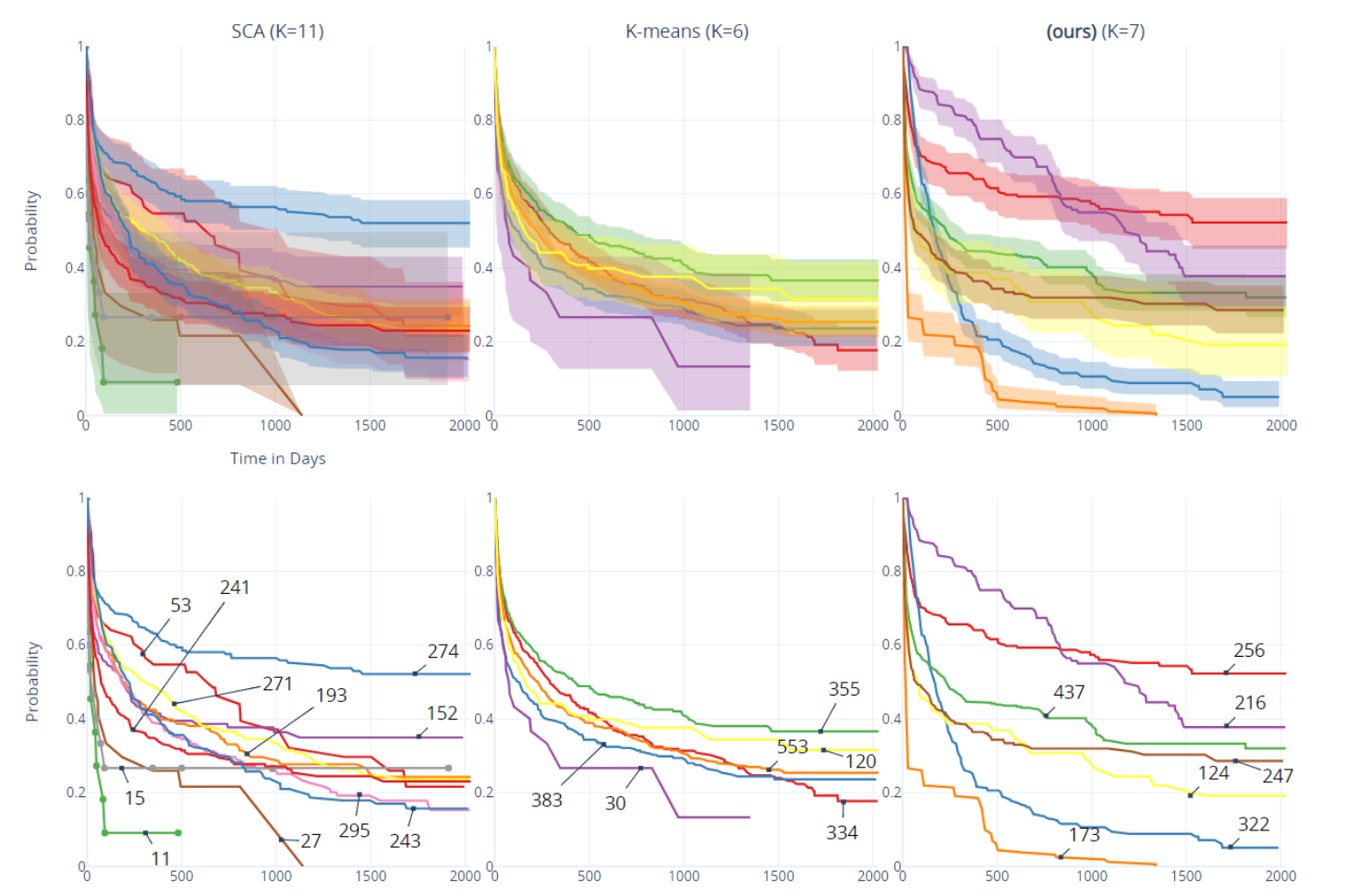}}
\caption{Test set's clusterization for the SUPPORT dataset returned by the models: SCA, K-means, and our proposal (\nameModel). The initial row shows the Kaplan-Meier of the cluster's subpopulations and the calculated confidence intervals. The row below shows the same survival functions, but now, without the confidence intervals and the number of data points inside each cluster.}
\label{fig:SUPPORT clusterization}
\end{center}
\vskip -0.2in
\end{figure*}

Figure \ref{fig:SUPPORT clusterization} shows the clusterization of the SUPPORT dataset's test set. It is clear that both SCA and \nameModel \: produce more varied and widespread clusters compared to traditional K-means. This is explained since both techniques utilize label information in the clustering process, in contrast to \textbf{K-means Survival}, which relies solely on feature data. A distinct advantage of \nameModel is its ability to generate more balanced-sized clusters that are also more dispersed than those created by SCA and K-means. This trend was consistent across various experiments.

It's worth mentioning that the \nameModel that has been trained can be analyzed to demonstrate its interpretability through the multinomial logistic regression structure and assumptions, as represented in Figure \ref{fig:graph model}. The SUPPORT dataset, depicted in Figure \ref{fig:SUPPORT clusterization}, is utilized to build a prognostic model that evaluates the survival of seriously ill hospitalized adults over a period of time  \cite{Knaus1995}. By analyzing the coefficients obtained from the multinomial logistic regression, it's possible to understand how the results shown on the right-hand side of the panel in Figure \ref{fig:SUPPORT clusterization}, marked as "ours," are linked to the characteristics of the patient. For example, the orange and blue clusters, which have a lower survival rate, are associated with metastatic cancer since their coefficients have a value of 4 compared to non-metastatic or no cancer, significantly increasing the log odds of being classified within these clusters. Similarly, it can be observed from the coefficients that the feature serum creatinine does not affect the log odds of patients being placed in any group; the coefficients for each cluster are all close to zero. This indicates that this feature does not impact survival, which can help the investigator gain insights into the problem.

\begin{table}[t]
\caption{Publicly accessible datasets used for the benchmark.}
\label{table:datasets}
\vskip 0.15in
\begin{center}
\begin{small}

\begin{tabular}{lcc}
\toprule
Dataset Name & Shape & Censoring ($ \frac{\sum_{i=1}^{N}\mathbb{I}\left(d_{i}=0\right)}{N}$)\\
\midrule
SUPPORT & $\left(8873, 14\right)$ & $31.9\%$\\
FLCHAIN & $\left(7874, 26\right)$ & $72.4\%$\\
METABRIC & $\left(1904, 9\right)$ & $42.0\%$\\
GBSG & $\left(2232, 7\right)$ & $43.2\%$\\
WHAS500 & $\left(500, 14\right)$ & $43.0\%$\\
\bottomrule
\end{tabular}

\end{small}
\end{center}
\vskip -0.1in
\end{table}

\section{Discussion}

% The clusterizations inferred by \nameModel reveal diverse population distributions with distinct survival profiles, making it highly applicable to our goal of discovering different groups. Additionally, most runs produce balanced clusters, and detailed plots can be found in the \nameAppendix. The model's simplicity and lightweight design enable it to perform competitively with other survival models. Furthermore, the structure and training of the model allow for the possibility of adapting \nameModel for other kinds of data by replacing the multinomial regression for other kinds of models having the same output.

The \nameModel model clusters survival data and provides a corresponding survival function for each point that can be used for prediction. Positive aspects of this model include: (i) a majority of runs result in balanced clusters; (ii) it possesses better predictive performance regarding the c-index metric when compared to algorithms that cluster across all datasets; (iii) it displays competitive performance when compared to predictive algorithms. Its performance is similar to GBSG and WHAS500 but falls into the second tier on METABRIC and FLCHAIN, being statistically similar to RSF and CoxPH. On the SUPPORT dataset, It ranks third alongside CoxPH in terms of performance.; (iv) it performed better in three out of the five datasets in the log-rank metric against the other clustering algorithms as shown in figure \ref{fig:logrank}.

While using our algorithm, it's important to note that there might be a higher variance in metrics between different runs with the same training data compared to algorithms that do not use Expectation Maximization in training. Therefore, it may be necessary to run multiple EM runs to select the best hyperparameter K or to find the most suitable trained model, as some degree of exploration is required.

It is worth noting that \textbf{K-means Survival} demonstrates competitive performance compared to SCA, as shown in Figures \ref{fig:time-dependent cindex} and \ref{fig:logrank}. However, it is possible that this performance gap could shrink when working with larger datasets, given that SCA leverages neural networks that have the potential to scale better in terms of performance.

Finally, there are some guiding principles to remember for proper model use. Firstly, even if a specific trained model acquires high values of C-index, it is advisable to visually check the clusterization on the training and validation sets before making a choice. Secondly, suppose there is a need to return the survival function for a data point. In that case, it is better to do full inference using equation \ref{eq:survival function} instead of just returning the survival function of its most probable cluster. When we calculate the c-index performance of \nameModel, this is the process we follow.

\section{Related Works}

Our work uses the traditional model-based approach for clustering, as outlined in \cite{bouveyron2019model}. The statistical literature includes works that utilize mixtures of parametric distributions for addressing the survival analysis problem, such as \cite{zeller_finite_2019, lachos_finite_2017, de_alencar_finite_2020, wang_model-based_2019}. However, our main objective is to create a mixture with nonparametric distributions. Also, models such as RankDeepSurv \cite{jing2019deep} or the weighted random forest \cite{utkin2019weighted} that cannot be immediately used to obtain clusters will not be directly compared.

In the machine learning literature, \citep{chapfuwa2020survival} provides a Bayesian nonparametric approach under the name Survival Cluster Analysis (SCA), using latent representation and distribution matching techniques. Our method's model structure differs from SCA as follows. Firstly, we incorporate features using multinomial logistic regression instead of a neural network. Secondly, for training, our model (\nameModel) utilizes the Expectation-Maximization (EM) algorithm rather than the stochastic gradient descent on minibatches employed by the Survival Cluster Analysis (SCA). Lastly, regarding the choice of the number of clusters, ours is determined by a hyperparameter, unlike the approach taken by SCA. 

\citep{mouli2019deep} introduced the DeepCLife model, which clusters data by optimizing pairwise distances using the log-rank score, a prevalent method in survival analysis. In contrast to DeepCLife's exclusive focus on clustering, our model concurrently optimizes both clustering and the prediction of time until an event, akin to the SCA model \cite{chapfuwa2020survival}. What further differentiates our approach is the incorporation of a latent space perspective. This enables a more nuanced understanding and representation of the data, enhancing both clustering and predictive accuracy, thus distinguishing it significantly from DeepCLife.

The VaDeSC algorithm, proposed by Manduchi et al. \cite{manduchi2021deep}, is a variational deep survival clustering model. It employs a Variational Auto-Encoder (VAE) regularized with a Gaussian mixture, creating a latent space that controls a survival density function, represented as a mixture of Weibull distributions. A key distinction between VaDeSC and our algorithm is its reliance on a parametric approach.

A new approach called SurvivalLVQ was very recently introduced for survival analysis \cite{de2024survivallvq}. This method uses Learning Vector Quantization (LVQ) to cluster survival data and shares some similarities with our goals, such as a focus on interpretability, the ability to return a survival function for each data point, and the use of the censored outcome label together with clusterization objectives in training. However, the main difference is that our approach is model-based, and as such, it is based on an explicit probability model for the full data set, inheriting all advantages of model-based approaches. It enables the use of standard statistical methodology and facilitates principled inference \cite{bouveyron2019model}. For example, doing model selection via the Bayesian Information Criterion (BIC) or estimating the uncertainty associated with the allocation of the $\boldsymbol{x}_{n}$’s cluster membership via the following equation, $r_{nk}=\mathrm{P}\left(z_{n}=k\mid t_{n}, d_{n}, x_{n}; \boldsymbol{\theta}^{(t)}\right)$:

\begin{equation}
{\text{Uncertainty}_n} = 1 - \max_{k=1,...,K} r_{nk}
\end{equation}

When we have a data point that lies halfway between two clusters, the uncertainty for that data point is expected to be high. Furthermore, outliers can be handled by including an extra component in the mixture model that represents them. The easiest way to do this is to add an additional component to the mixture model in equation \ref{eq:model density} that has a uniform distribution in the data region. Here, $\tau_{0}$ denotes the expected proportion of outliers in the data and $V$ denotes the volume of the data region:

\begin{equation}
\label{eq:model density plus constant}
f\left({t}_{i}^{*} \mid \boldsymbol{x}_{i}\right)=\frac{\tau_{0}}{V} + \sum_{k=1}^{K} \tau_{k}\left(\boldsymbol{x}_{i}\right) f\left({t}_{i}^{*} \mid \theta_{k}\right)
\end{equation}

Our model is an extension of the one proposed by \citep{bordes2016stochastic}, employing a similar non-parametric approach for clustering time labels. However, we have introduced several modifications. Notably, we utilize the features to model the mixing proportions, enabling our model to identify more than two clusters. Additionally, we have made other modifications to improve its functionality. We also conduct a comparative analysis to evaluate the predictive performance of our algorithm against other established models.

\section{Conclusion}

The present paper proposes a new survival model that clusters while also predicting survival functions. The algorithm successfully adapts the non-parametric model-based clusterization framework for right-censored time-to-event data. Through experimentation across five public datasets, we found that (i) most of the runs generate well-balanced clusters, (ii) the algorithm has superior predictive performance in terms of the c-index metric when compared to clustering algorithms, and it can compete with purely predictive survival models; (iii) it outperformed other clusterization algorithms in three out of five datasets when evaluated using the log-rank metric; (iv) the architecture of the model can be easily adapted to accommodate features with exploitable structures. For example, practitioners can use this tool to identify evidence of heterogeneous treatment effects. 

The model's code is publicly accessible and available at \href{https://github.com/buginga/SurvMixClust}{https://github.com/buginga/SurvMixClust}.

\bibliography{survival_cluster_Buginga}
\bibliographystyle{icml2020}

\appendix

\section{Experimental Setup}

The datasets were split into training, validation, and testing sets with 60\%, 20\%, and 20\% division, stratified by censoring rate. Then, a total of 20 different splits helped to generate 20 performance metrics samples. When a hyperparameter search occurred, the testing and validation set were combined, and a 3-fold cross-validation was applied, returning the best hyperparameter. This same chosen hyperparameter was selected to retrain with the whole combined training plus validation, then calculating the needed metrics.  

\section{Visualizing the survival functions}

\begin{figure*}[ht]
\vskip 0.2in
\begin{center}
\centerline{\includegraphics[width=\textwidth]{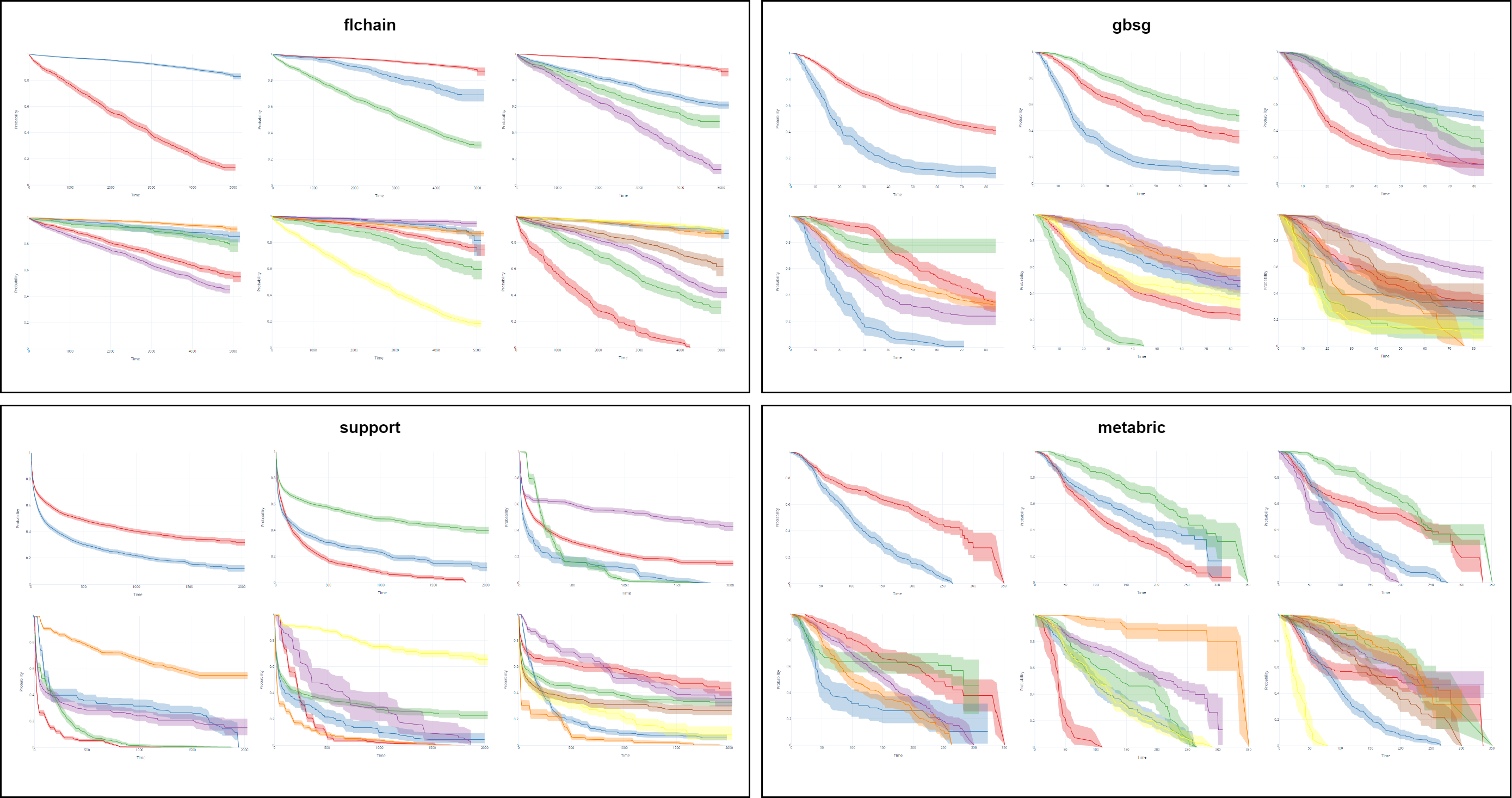}}
\caption{Inferred clusterizations generated by \nameModel. Each card corresponds to a dataset. A randomly selected trained model for each number of clusters is used to cluster the test set. The survival function of these populations, via Kaplan-Meier, is exhibited inside each card.}
\label{fig:clusters results}
\end{center}
\vskip -0.2in
\end{figure*}

Figure \ref{fig:clusters results} displays the survival functions of the clusterized populations generated by \nameModel. Each card corresponds to a dataset inside the paper's main body. A randomly selected trained model for each number of clusters (2, 3, 4, 5, 6, 7) is used to cluster the test set. The survival function of these grouped populations, via Kaplan-Meier, is exhibited inside each card. 

The main text mentions that most inferred clusterization displays good qualitative distribution of heterogeneous populations, discovering different survival profiles. A further point is the balanced character of each cluster; no cluster has too few members, as can be immediately seen by the size of the Kaplan-Meier's confidence interval.

\section{Hyperparameters}

\begin{table}[ht]
\label{tab:params_table}
\resizebox{\columnwidth}{!}{\begin{tabular}{|c|l|}
\hline
\textbf{Model}   & \multicolumn{1}{c|}{\textbf{Hyperparameters}}                                                                                                                                                                                                                                                                                                                                                                                                 \\ \hline
RSF              & \begin{tabular}[c]{@{}l@{}}max tree depth: {[}3, 5, 8{]},\\ min \# of samples required to be at a leaf node: {[}20, 50, 150{]},\\ number of trees: {[}50, 100, 200{]}\end{tabular}                                                                                                                                                                                                                                                        \\ \hline
CoxPH            & \begin{tabular}[c]{@{}l@{}}l2 regularization factor: {[}0.0001, 0.01, 1{]},\\ lr: {[}1, 0.5, 0.01{]}\end{tabular}                                                                                                                                                                                                                                                                                                                             \\ \hline
K-means Survival & number of clusters: {[}2, 3, 4, 5, 6, 7{]}                                                                                                                                                                                                                                                                                                                                                                                                    \\ \hline
\textbf{(ours)}  & number of clusters: {[}2, 3, 4, 5, 6, 7{]}                                                                                                                                                                                                                                                                                                                                                                                                    \\ \hline
Logistic Hazard  & \begin{tabular}[c]{@{}l@{}}neural network architecture: {[}{[}1{]}, {[}2{]}, {[}3{]}, {[}5{]}, {[}8{]},\\                                                  {[}2, 2{]}, {[}3, 3{]}, {[}5, 5{]}, {[}8, 8{]},\\                                                  {[}2, 2, 2{]}, {[}3, 3, 3{]}, {[}5, 5, 5{]}, {[}8, 8, 8{]}{]},\\ dropout: {[}0, 0.2, 0.5, 0.8{]},\\ number of divisions of the output time axis: {[}10, 50, 100{]}\end{tabular} \\ \hline
SCA              & default from the paper's code                                                                                                                                                                                                                                                                                                                                                                                                                 \\ \hline
\end{tabular}}
\end{table}

\section{Mathematical Derivations}
\label{derivations}
We will use the notation in Table \ref{table:notation} and the same assumptions to explain how we derived equations \ref{eq:trick1}, \ref{eq:trick2}, \ref{eq:LL} and \ref{eq:expectation}. To begin with, let's consider equation \ref{eq:trick1}. We need to recall the definition of random variables $T=\text{min}\{T^{*},C^{*}\}$ and $D=\mathds{1}\{T^{*}\leq C^{*}\}$. These probability events can be rewritten in a better format for our purposes:
\resizebox{.9\columnwidth}{!}{
\begin{minipage}{\columnwidth}

\begin{equation}
\begin{aligned}
&\mathrm{P}(T=t, D=d)=\\
&=\mathrm{P}\left(T^{*}=t, C^{*} \geq t\right)^{\mathbb{I}\left(d=1\right)} \mathrm{P}\left(T^{*}>t, C^{*}=t\right)^{\mathbb{I}\left(d=0\right)} \\
&=\left[\mathrm{P}\left(T^{*}=t\right) \mathrm{P}\left(C^{*} \geq t\right)\right]^{\mathbb{I}\left(d=1\right)}\left[\mathrm{P}\left(T^{*}>t\right) \mathrm{P}\left(C^{*}=t\right)\right]^{\mathbb{I}\left(d=0\right)} \\
&=\left[f(t)\left(S_{C^{*}}(t)+f_{C^{*}}(t)\right)\right]^{\mathbb{I}\left(d=1\right)}\left[S(t) f_{C^{*}}(t)\right]^{\mathbb{I}\left(d=0\right)} \\
&=\left[f(t)^{\mathbb{I}\left(d=1\right)} S(t)^{\mathbb{I}\left(d=0\right)}\right]\left[f_{C^{*}}(t)^{\mathbb{I}\left(d=0\right)}\left(S_{C^{*}}(t)+f_{C^{*}}(t)\right)^{\mathbb{I}\left(d=1\right)}\right]
\end{aligned}
\end{equation}

\end{minipage}
}

Equation \ref{eq:expectation} for $r_{nk}^{(t)}$ is derived using $h(t)=\left[f_{C^{*}}(t)^{\mathbb{I}\left(d=0\right)}\left(S_{C^{*}}(t)+f_{C^{*}}(t)\right)^{\mathbb{I}\left(d=1\right)}\right]$ and $h(t_{n}|z_{n}=l)=h(t_{n})$:

\resizebox{.9\columnwidth}{!}{
\begin{minipage}{\columnwidth}

\begin{equation}
\begin{aligned}
r_{nk}^{(t)}&=\mathrm{P}\left(z_{n}=k\mid t_{n}, d_{n}, x_{n}; \boldsymbol{\theta}^{(t)}\right)\\
&=\frac{\mathrm{P}\left(t_{n}, d_{n}\mid z_{n}=k; \boldsymbol{\theta}^{(t)}\right)\mathrm{P}\left(z_{n}=k \mid x_{n}; \boldsymbol{\theta}^{(t)}\right)}{\sum_{l=1}^{K} \mathrm{P}\left(t_{n}, d_{n}\mid z_{n}=l; \boldsymbol{\theta}^{(t)}\right)\mathrm{P}\left(z_{n}=l \mid x_{n}; \boldsymbol{\theta}^{(t)}\right)} \\
&=\frac{\mathrm{P}\left(t_{n}, d_{n}\mid z_{n}=k; \boldsymbol{\theta}^{(t)}\right)\tau_{k}^{(t)}\left(\boldsymbol{x}_{i}\right)}{\sum_{l=1}^{K} \mathrm{P}\left(t_{n}, d_{n}\mid z_{n}=l; \boldsymbol{\theta}^{(t)}\right)\tau_{l}^{(t)}\left(\boldsymbol{x}_{i}\right)}\\
&=\frac{\mathrm{P}\left(t_{n}, d_{n}\mid z_{n}=k; \boldsymbol{\theta}^{(t)}\right)\tau_{k}^{(t)}\left(\boldsymbol{x}_{i}\right)}{\sum_{l=1}^{K} \mathrm{P}\left(t_{n}, d_{n}\mid z_{n}=l; \boldsymbol{\theta}^{(t)}\right)\tau_{l}^{(t)}\left(\boldsymbol{x}_{i}\right)}\\
&=\frac{{f\left(t_{n} \mid \theta_{k}\right)}^{\mathbb{I}\left(d_{n}=1\right)}{S\left(t_{n} \mid \theta_{k}\right)}^{\mathbb{I}\left(d_{n}=0\right)}h(t_{n}|z_{n}=k)\tau_{k}^{(t)}\left(\boldsymbol{x}_{i}\right)}{\sum_{l=1}^{K} {f\left(t_{n} \mid \theta_{l}\right)}^{\mathbb{I}\left(d_{n}=1\right)}{S\left(t_{n} \mid \theta_{l}\right)}^{\mathbb{I}\left(d_{n}=0\right)}h(t_{n}|z_{n}=l)\tau_{l}^{(t)}\left(\boldsymbol{x}_{i}\right)}\\
&=\frac{{f\left(t_{n} \mid \theta_{k}\right)}^{\mathbb{I}\left(d_{n}=1\right)}{S\left(t_{n} \mid \theta_{k}\right)}^{\mathbb{I}\left(d_{n}=0\right)}h(t_{n})\tau_{k}^{(t)}\left(\boldsymbol{x}_{i}\right)}{\sum_{l=1}^{K} {f\left(t_{n} \mid \theta_{l}\right)}^{\mathbb{I}\left(d_{n}=1\right)}{S\left(t_{n} \mid \theta_{l}\right)}^{\mathbb{I}\left(d_{n}=0\right)}h(t_{n})\tau_{l}^{(t)}\left(\boldsymbol{x}_{i}\right)}\\
&=\frac{{f\left(t_{n} \mid \theta_{k}\right)}^{\mathbb{I}\left(d_{n}=1\right)}{S\left(t_{n} \mid \theta_{k}\right)}^{\mathbb{I}\left(d_{n}=0\right)}\cancel{h(t_{n})}\tau_{k}^{(t)}\left(\boldsymbol{x}_{i}\right)}{\sum_{l=1}^{K} {f\left(t_{n} \mid \theta_{l}\right)}^{\mathbb{I}\left(d_{n}=1\right)}{S\left(t_{n} \mid \theta_{l}\right)}^{\mathbb{I}\left(d_{n}=0\right)}\cancel{h(t_{n})}\tau_{l}^{(t)}\left(\boldsymbol{x}_{i}\right)}\\
&=\frac{\left[{f\left(t_{n} \mid \theta_{k}\right)}^{\mathbb{I}\left(d_{n}=1\right)}{S\left(t_{n} \mid \theta_{k}\right)}^{\mathbb{I}\left(d_{n}=0\right)}\right]\tau_{k}^{(t)}\left(\boldsymbol{x}_{i}\right)}{\sum_{l=1}^{K} \left[{f\left(t_{n} \mid \theta_{l}\right)}^{\mathbb{I}\left(d_{n}=1\right)}{S\left(t_{n} \mid \theta_{l}\right)}^{\mathbb{I}\left(d_{n}=0\right)}\right]\tau_{l}^{(t)}\left(\boldsymbol{x}_{i}\right)}\\
\end{aligned}
\end{equation}

\end{minipage}
}

Equation \ref{eq:LL} is a direct application of \ref{eq:trick2}:

\resizebox{.9\columnwidth}{!}{
\begin{minipage}{\columnwidth}

\begin{equation}
\begin{aligned}
& L L^{t}(\boldsymbol{\theta})=\\
& \sum_{n} \mathbb{E}_{q_{n}^{t}\left({z}_{n}\right)}\left[\log \mathrm{P}(T=t_{n}, D=d_{n}, Z=z_{n} \mid X=x_{n}; \boldsymbol{\theta}) \right]\\
&=\sum_{n} \mathbb{E}_{q_{n}^{t}\left({z}_{n}\right)}\left[\log \mathrm{P}_{\boldsymbol{\theta}}(Z=z_{n}\mid X=x_{n}){f\left(t_{n} \mid \theta_{z_{n}}^{(t)}\right)}^{\mathbb{I}\left(d_{n}=1\right)}\right.\cdot\\
&\qquad \qquad \qquad \qquad \qquad \qquad \cdot \left. {S\left(t_{n} \mid \theta_{z_{n}}^{(t)}\right)}^{\mathbb{I}\left(d_{n}=0\right)}h(t_{n}) \right]\\
&=\sum_{n} \mathbb{E}_{q}\left[\log \tau_{z_{n}}\left(\boldsymbol{x}_{n}\right) {f\left(t_{n} \mid \theta_{z_{n}}^{(t)}\right)}^{\mathbb{I}\left(d_{n}=1\right)}\right.\cdot\\
&\qquad \qquad \qquad \qquad \qquad \qquad \cdot \left. {S\left(t_{n} \mid \theta_{z_{n}}^{(t)}\right)}^{\mathbb{I}\left(d_{n}=0\right)}h(t_{n}) \right]\\
&=\sum_{n} \mathbb{E}_{q}\left[\log \prod_{k} \left(\tau_{k}\left(\boldsymbol{x}_{n}\right) {f\left(t_{n} \mid \theta_{k}^{(t)}\right)}^{\mathbb{I}\left(d_{n}=1\right)}\cdot\right. \right.\\
&\qquad \qquad \qquad \qquad \qquad \qquad \cdot\left. {\left.{S\left(t_{n} \mid \theta_{k}^{(t)}\right)}^{\mathbb{I}\left(d_{n}=0\right)}h(t_{n})\right)}^{z_{nk}} \right]\\
&= \sum_{n} \sum_{k} \mathbb{E}\left[z_{n k}\right]\log \tau_{k}\left(\boldsymbol{x}_{n}\right) + \\
&\qquad + \sum_{n} \sum_{k} \mathbb{E}\left[z_{n k}\right]\log {f\left(t_{n} \mid \theta_{k}^{(t)}\right)}^{\mathbb{I}\left(d_{n}=1\right)}{S\left(t_{n} \mid \theta_{k}^{(t)}\right)}^{\mathbb{I}\left(d_{n}=0\right)} +\\
& \qquad + \sum_{n} \sum_{k} \mathbb{E}\left[z_{n k}\right] \log \left(h(t_{n})\right)\\
\end{aligned}
\end{equation}

\end{minipage}
}

When searching for the value of $q^{(l)}_{i}$, the necessary terms can be found using a straightforward relation. Notice that the denominator is not dependent on $k$, so it can be factored out of the $\arg \max$.

\resizebox{.9\columnwidth}{!}{
\begin{minipage}{\columnwidth}

\begin{equation}
\begin{aligned}
&q^{(t)}_{i}=\arg \max _{k} \left( r_{ik}^{(t)}\right)\\
&q^{(t)}_{i}=\\
&\arg \max _{k}\frac{\left[{f\left(t_{i} \mid \theta_{k}^{(t)}\right)}^{\mathbb{I}\left(d=1\right)}{S\left(t_{i} \mid \theta_{k}^{(t)}\right)}^{\mathbb{I}\left(d=0\right)}\right]\tau_{k}^{(t)}\left(\boldsymbol{x}_{i}\right)}{\sum_{l=1}^{K} \left[{f\left(t_{i} \mid \theta_{l}^{(t)}\right)}^{\mathbb{I}\left(d=1\right)}{S\left(t_{i} \mid \theta_{l}^{(t)}\right)}^{\mathbb{I}\left(d=0\right)}\right]\tau_{l}^{(t)}\left(\boldsymbol{x}_{i}\right)}\\
&q^{(t)}_{i}=\arg \max _{k}\left(\left[{f\left(t_{i} \mid \theta_{k}^{(t)}\right)}^{\mathbb{I}\left(d=1\right)}{S\left(t_{i} \mid \theta_{k}^{(t)}\right)}^{\mathbb{I}\left(d=0\right)}\right]\tau_{k}^{(t)}\left(\boldsymbol{x}_{i}\right)\right)\\
\end{aligned}
\end{equation}

\end{minipage}
}

%%%%%%%%%%%%%%%%%%%%%%%%%%%%%%%%%%%%%%%%%%%%%%%%%%%%%%%%%%%%%%%%%%%%%%%%%%%%%%%
%%%%%%%%%%%%%%%%%%%%%%%%%%%%%%%%%%%%%%%%%%%%%%%%%%%%%%%%%%%%%%%%%%%%%%%%%%%%%%%

\end{document}